\pdfoutput=1

\documentclass[11pt]{article}

\usepackage{ACL2023}

\usepackage{times}
\usepackage{latexsym}
\usepackage{booktabs}
\usepackage{todonotes}
\usepackage{pgfplots} 
\usepackage{gnuplottex}

\usepackage[T1]{fontenc}

\usepackage[utf8]{inputenc}

\usepackage{microtype}

\usepackage{inconsolata}

\usepackage{subcaption}

\title{Application-Agnostic Language Modeling for On-Device ASR}

\author{Markus Nu{\ss}baum-Thom \qquad Lyan Verwimp \qquad Youssef Oualil \\
  Apple \\
  \texttt{\{mnussbaumthom,lverwimp,youalil\}@apple.com}\\
  accepted at ACL 2023 industry track} 

\begin{document}
\maketitle
\begin{abstract}
On-device automatic speech recognition systems face several challenges compared to server-based systems.
They have to meet stricter constraints in terms of speed, disk size and memory while maintaining the same accuracy.
Often they have to serve several applications with different distributions at once, such as communicating with a virtual assistant and speech-to-text.
The simplest solution to serve multiple applications is to build application-specific (language) models, but this leads to an  increase in memory. 
Therefore, we explore different data- and architecture-driven language modeling approaches to build a single application-agnostic model. We propose two novel feed-forward architectures that find an optimal trade off between different on-device constraints.
In comparison to the application-specific solution, one of our novel approaches reduces the disk size by half, while maintaining speed and accuracy of the original model.

\end{abstract}

\section{Introduction}
\label{sec:intro}

On-device Automatic Speech Recognition (ASR) is subject to several constraints: it should return accurate results in a reasonable time frame without consuming too much memory and disk space. 
State-of-the-art research often is accuracy focused, while resource-constrained applications also need to take care of performance and size. Finding an architecture that reaches all constraints is not trivial.

Another challenge is that ASR systems often 
serve a large variety of requests. 
ASR systems
can serve an on-device Virtual Assistant (VA) but also allow dictated messages, notes, e-mails, etc. -- we refer to the latter application as Speech-to-text (STT).
Typical VA requests are knowledge-driven questions such as~\textit{``how old is Barack Obama?''} or commands, e.g.~\textit{``play some Lady Gaga music''}.
\
STT requests are 
longer and of a different nature than typical VA requests. 
The solution that yields the best accuracy for both VA and STT is to train separate models for each application, but additional model size is prohibitive.
We aim to develop a single model instead. 

In this paper, we focus on a Neural Network Language Model (NNLM) in the ASR system. Our baseline is a Fixed-size Ordinally-Forgetting Encoding (FOFE) feed-forward NNLM~\cite{zhang}.
In ASR, the search space can easily increase so we have to limit the context length used in decoding to reach an acceptable latency and lower memory. Given this short context length, we find that the FOFE feed-forward LM is competitive to the Transformer~\cite{vaswani} in terms of accuracy and better in terms of latency.
\citet{iriethesis} has also shown that Transformers are less robust to short context lengths. 

To build a single Application-Agnostic (AA) NNLM, we developed a method to optimally sample training data.
We sample data from different sources, e.g. anonymized and randomly sampled user requests from opted-in users for VA and STT and artificial requests spanning many different domains that focus on improving the tail of the distribution.
The data-driven approach tries to find the optimal balance between the application-specific data sources by creating a balanced development set and distributing the sampling weights based on the importance of each data source and each application on that development set.

Training a single FOFE NNLM on the combined dataset can lead to accuracy degradations, even with a larger model or longer training. 
We explore two extensions to the baseline FOFE NNLM:
firstly, a Mixture FOFE NNLM~\cite{youalil,IrieKNL18} which is composed of an ensemble of parallel sub-networks and a mixture sub-network generating normalized probabilities across all sub-networks. 
These mixture weights are used to compute a weighted average of the ensemble before the softmax output. 
The second extension is an Application-Dependent (AD) FOFE NNLM that has different sub-networks for each application. 
At training time, data and gradients are (back-)propagated only through the corresponding sub-network belonging to an application. At inference time, the way the user invokes ASR tells us which application is needed (wake phrase = VA, microphone button = STT) and only the sub-network belonging to the active application is used.
Both approaches are able to match or outperform the application-specific model.
While the accuracy of the mixture NNLM is slightly better than the AD-NNLM the situation is reversed in terms of speed.

The contributions of this paper are as follows:
\begin{itemize}
 \item We propose a method to optimally combine application-specific data sources to train an application-agnostic LM in Section~\ref{sec:data}.
 \item We propose two novel FOFE-based neural LMs in Section~\ref{sec:models} that each match the accuracy of two application-specific language models.
 \item In Section~\ref{sec:results} we compare the novel NNLMs accuracy and speed against the baseline FOFE and state-of-art Transformer models. We do this for three different languages - US English, German and Mandarin Chinese - and three types of test sets (see Section~\ref{sec:setup} for more information).
\end{itemize}



\section{Related work}
\label{sec:related}
We start by discussing related work on modeling several domains/tasks at once.
Many pattern recognition tasks are imbalanced since data from different categories do not occur at the same frequency. Therefore, the less frequent categories are not well represented in the training data~\cite{AnandMMR93,Johnson}, which results in a sub-optimal model.
Data-driven approaches to deal with the data imbalance include under- and over-sampling~\cite{HulseKN07}. 
Refinements of these methods select data more intelligently \cite{KubatM97,chawla2002smote,mani2003knn,barandela2004imbalanced}.

Others approaches modify the training and/or model architecture.
Curriculum Learning~\cite{BengioLCW09,ShiLJ15} emphasizes data by fine-tuning  towards the corpus consumed by the end of training. \citet{SmithWSWB20} experiment with multi-task learning, data augmentation and a classifier combined with single-task models to appropriately model several skills in a conversation agent.
Balancing through interleaved sampling of different corpora was investigated in \cite{xing2022} as well as model-based approaches like multi-task and weighted learning, which allows the model to self-control the impact of different corpora. Other ways to increase the modeling power are using a Mixture of Experts~\cite{shazeer,yanqizhou2022} or ensemble networks~\cite{youalil,IrieKNL18,ganaie}.

 The choice of architecture for language modeling has also been a recurrent topic of research.
Early neural LMs use feed-forward layers~\cite{schwenk,bengio}. \citet{mikolov} introduced recurrent neural LMs that can in principle use unlimited history. These networks are trained with back-propagation through time which `unrolls' the network in time for gradient computation, but this leads to vanishing gradients~\cite{bengio2,pascanu}, essentially limiting the history that can be learned from. Gated recurrent architectures~\cite{sundermeyer,Cho} mitigate this problem.

Recent extensions of the feed-forward architecture have been proposed that alleviate different disadvantages. \citet{zhang} proposed a FOFE, which represents a sequence of words as a vector with fixed length that captures the word order. They show that feed-forward networks with FOFE encoding outperform recurrent models in language modeling. 
The most widely-used architecture in recent years, is the Transformer~\cite{vaswani} that combines feed-forward layers with multi-head attention, residual connections, and layer normalization~\cite{ba}. It has been successfully applied to ASR, see e.g.~\cite{irie,beck}. In this paper, we compare FOFE feed-forward LMs with Transformer LMs and two extensions of the base FOFE feed-forward LMs.

\section{Data balancing}
\label{sec:data}

The ASR system in this paper serves two applications, VA and STT, for which we observe very different linguistic patterns. To demonstrate these differences, we calculate statistics on two English development sets.
Each data set contains 23k anonymized queries and is randomly sampled from real user data similarly to the test sets described in Section~\ref{sec:setup}.

VA requests are typically shorter than STT requests. In Figure~\ref{fig:num_words}, we plot the number of queries (on a logarithmic scale) that have \textit{x} number of words for both data sets. For example, in the VA dev set there are 9968 requests with only two words (238 requests consist of only \textit{``{\textless}wakeword\_1{\textgreater} {\textless}wakeword\_2{\textgreater}''}), while the STT test set contains 1327 requests with two words. If we define a request of 30 or more words as a `long' request, we see that the STT test has 2030 long requests while VA has only 21 long requests.



\begin{figure}[t]
\centering
\resizebox{7.5cm}{6.5cm}{
  \begin{tikzpicture}
\begin{semilogyaxis}[
    enlargelimits=false,
    xlabel={\# of words per query},
    ylabel={\# of queries with $x$ \# of words (log)},
    legend style={at={(0.8,0.9)},anchor=north,legend columns=-1},
]
\addplot+[only marks, color=blue, mark=*, mark size=1.5pt] table {va.dat};
\addlegendentry{VA};
\addplot+[only marks, color=red, mark=triangle*, mark size=1.5pt] table {stt.dat};
\addlegendentry{STT};
\end{semilogyaxis}
\end{tikzpicture}
}
\setlength{\abovecaptionskip}{0cm}
\setlength{\belowcaptionskip}{-10pt}
\caption{Number of queries (on the y-axis in log scale) with \textit{x} number of words (on the x-axis) in the English VA and STT dev sets.}
\label{fig:num_words}
\end{figure}
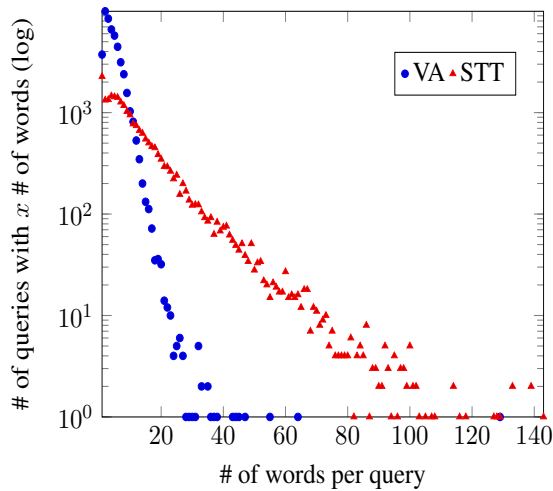

\begin{figure}
\centering
\begin{tikzpicture}[scale=0.85]
\begin{axis} [
    xbar interval=1.0,
    xmin = 0,
    xmax = 11000,
    xticklabel style={
        /pgf/number format/fixed,
        /pgf/number format/precision=0
    },
    scaled x ticks=false,
    symbolic y coords={in,that,it,.,and,you,I,a,for,what,text,to,is,the,wake1,wake2,call},
    ytick={call,wake2,wake1,the,is,to,text,what,for,a,I,you,and,.,it,that,in},
    legend style={
            at={(0.82,0.17)},
            anchor=north,
            legend columns=-1,
    },
    legend image code/.code={
        \draw [#1] (0cm,-0.1cm) rectangle (0.2cm,0.25cm); },
    ]
    \addplot[fill=blue,bar width=3pt] coordinates {
        (6800,call)
        (6049,wake2)
        (4834,wake1)
        (2278,the)
        (1703,is)
        (690,to)
        (1486,text)
        (1474,what)
        (1438,for)
        (1374,a)
        (690,I)
        (1366,you)
        (353,and)
        (104,.)
        (492,it)
        (155,that)
        (1082,in)
    };
    \addlegendentry{VA}
    \addplot[draw=red,fill=red,bar width=3pt] coordinates {
        (582,call)
        (10,wake2)
        (378,wake1)
        (8199,the)
        (2627,is)
        (9134,to)
        (268,text)
        (1362,what)
        (2863,for)
        (4861,a)
        (10304,I)
        (9686,you)
        (6350,and)
        (5822,.)
        (3628,it)
        (3332,that)
        (2898,in)
    };
    \addlegendentry{STT}
\end{axis}
\end{tikzpicture}
\caption{Counts of the union of the 10 most frequent words in both English VA and STT dev sets. ``wake2'' and ``wake1'' refer to \textit{``{\textless}wakeword\_2{\textgreater}} and \textit{``{\textless}wakeword\_1{\textgreater}}.}
\label{fig:top25}
\end{figure}
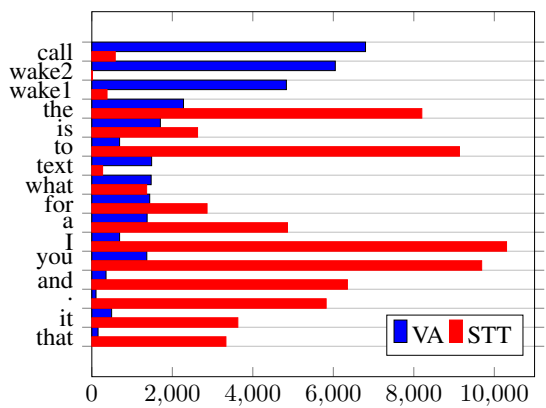

Secondly, the content and style of the requests varies between the two applications. Figure~\ref{fig:top25} plots the union of the top 10 most frequent words in each data set  -- ordered by the frequency in the VA dev set. 
Notice that we allow the user to also dictate punctuation marks, hence the presence of the dot in the list of words. 
It is clear from this distribution that VA queries are often questions (\textit{what}) or commands (\textit{call, text}) while STT queries are often messages from the perspective of the user (\textit{I, you}) who wants to make their message more readable with punctuation marks. 


Because of the different linguistic nature of these two applications, balancing the NNLM training data has a large impact on the quality of the model. 
A common strategy to determine NNLM sampling weights for each application is to train individual $n$-gram LMs on each data source and choose relevance weights based on the optimal linear interpolation weights  on a development set~\cite{RajuFTLR19}.
In our setup, the sampling weights for the application-specific text sources are derived from the count merging weights~\cite{bacchiani,hsu,pusateri} instead of a linear combination.

We propose a balancing scheme to derive sampling weights for $I$ text sources that benefit both applications.
We create a balanced development set containing approximately the same amount of VA and STT data.
Let $\alpha_{1}, \ldots, \alpha_{I} \in [0, 1]$ be the sampling weights such that $\sum_{i=1}^I \alpha_i = 1$ and $\rho(i) \in \{D, A\}$ indicating if the text source belongs to STT or VA.
The redistribution probability masses $\beta_D$ and $\beta_A$  for STT and VA respectively are calculated to serve the joint application.
These probability masses are determined by the optimal weights that minimize the perplexity of the linear Application-Specific (AS) language model  combination on the balanced development set.
The application-specific probability mass allocated by each application can be formalized as:
\begin{displaymath}
    \overline{\alpha}_{D} := \sum_{i, \rho(i) = D} \alpha_i \qquad \textnormal{and} \qquad
    \overline{\alpha}_{A} := \sum_{i, \rho(i) = A} \alpha_i.
\end{displaymath}

\noindent Now consider the ratio between the redistribution and application-specific probability mass:
\begin{displaymath}
    \gamma_{A} := \frac{\beta_{A}}{\overline{\alpha}_A} \qquad \textnormal{and} \qquad \gamma_{D} := \frac{\beta_{D}}{\overline{\alpha}_D}.
\end{displaymath}

\noindent These ratios determine the scaling of the original sampling weights to achieve balancing.
Balanced sampling weights are then determined by a re-normalization of the scaled sampling weights:
\begin{displaymath}
    \lambda_{i} := \frac{\displaystyle\gamma_{\rho(i)}\alpha_i}{\displaystyle\sum_j \gamma_{\rho(j)} \alpha_j}, \qquad i=1, \ldots, I.
\end{displaymath}

\noindent The heldout and training set for NNLM training is then randomly sampled from the text sources according to the balanced sampling weights.


\begin{figure*}[!htb]
\centering
\begin{subfigure}[b]{0.3\textwidth}
    \centering
    \includegraphics[width=0.6\textwidth]{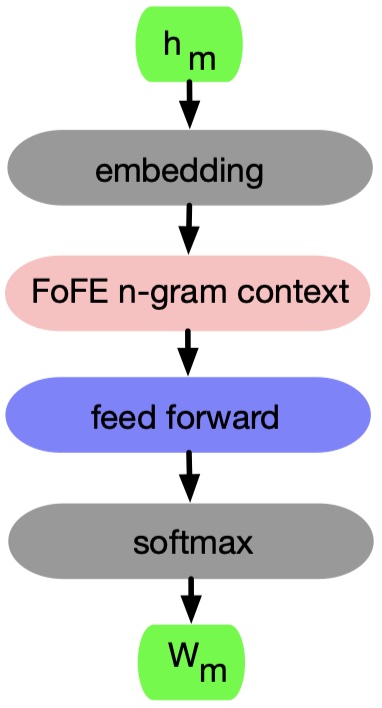}
    \caption{Base.}
    \label{fig:fofe}
\end{subfigure}
\hfill
\begin{subfigure}[b]{0.3\textwidth}
    \centering
    \includegraphics[width=0.9\textwidth]{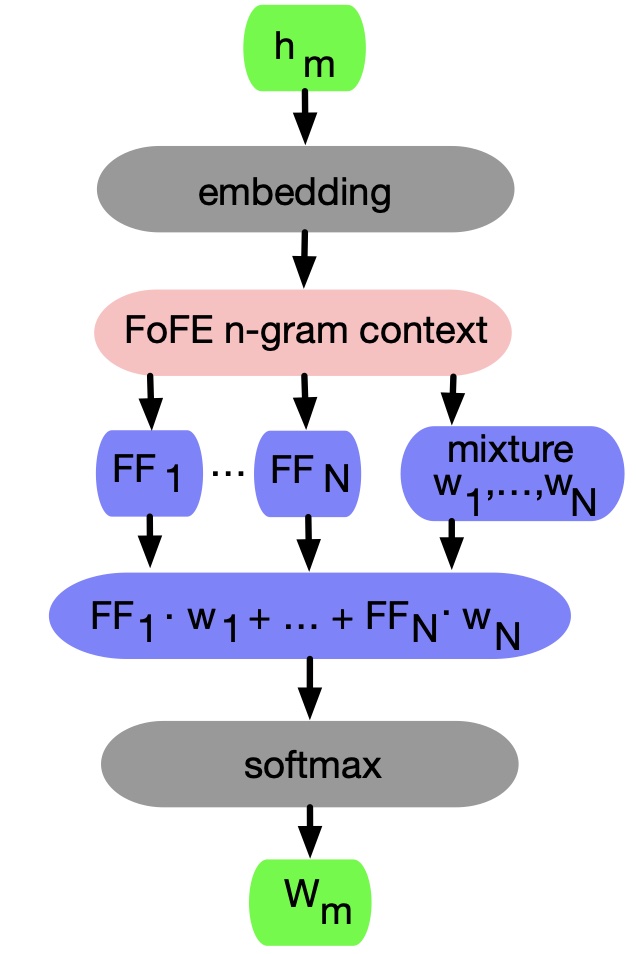}
    \caption{Mixture}
    \label{fig:fofe_mixture}
\end{subfigure}
\hfill
\begin{subfigure}[b]{0.3\textwidth}
\centering
\includegraphics[width=0.64\textwidth]{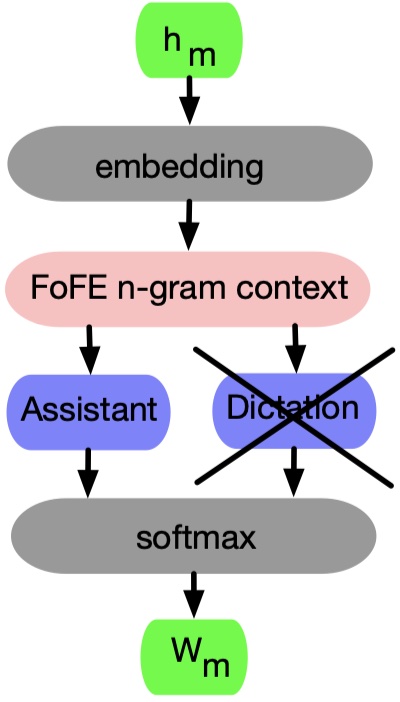}
\caption{Application-dependent.}
\label{fig:fofe_application_dependent}
\end{subfigure}
\caption{Let $w_m$ and history $h_m$ be the word and history at position $m$.}
\end{figure*}

\section{Application-Agnostic and Application-Dependent FOFE NNLMs}
\label{sec:models}
In this section three different types of NNLM architectures are introduced for on-device ASR. 
In the following let $w_1^N := w_1, \ldots, w_N$ be a word sequence.
All NNLM architectures considered here follow a similar scheme.  In each architecture a word embedding is followed by a FOFE layer~\cite{zhang}. Let $\alpha > 0$ be the forgetting factor of the FOFE layer and $e_m$ be the word embedding of word $w_m$ then $z_m := z_{m-1} + \alpha \cdot e_{m}$
generates the FOFE encoding. Afterwards an $n$-gram context of the FOFE encoding is generated by concatenating $n$ subsequent FOFE encodings for each position: $z_{m-n+1}, \ldots, z_{m}$.
Next, this context is flattened and passed to the hidden layers.

The baseline FOFE NNLM shown in Figure~\ref{fig:fofe} applies a stack of feed-forward layers to the flattened FOFE $n$-gram context. The output of the last feed-forward layer is fed to a projection layer for dimension reduction before the final softmax layer.
This  architecture is used for the 
AS-NNLM, where each application has its own NNLM, as well as for the Application-Agnostic (AA) NNLM, which is trained on balanced data for both applications. 

Figure~\ref{fig:fofe_mixture} shows the mixture NNLM, which has $M$ parallel sub-networks and a mixture sub-network. Each sub-network is a stack of feed-forward layers. The mixture sub-network is also a stack of feed-forward layers which finish with a softmax output of dimension $M$ to produce mixture weights for each of the parallel sub-networks, similarly to~\cite{youalil,IrieKNL18,yanqizhou2022} except that the mixture combines FOFE networks.
The subsequent layer averages the output of all parallel sub-networks scaled by the corresponding  weights of the mixture sub-network softmax output.  

Figure~\ref{fig:fofe_application_dependent} shows the Application-Dependent NNLM (AD-NNLM). This architecture uses the application information to train a NNLM in a multi-task style. This NNLM has a separate sub-network and softmax output biases for each application. For training we follow a multi-task approach. The information of the application for each data sample is known and used to select the sub-network and softmax output bias corresponding to the application and only back-propagate through a part of the NNLM. At inference time, data are forwarded through the corresponding sub-network and softmax output bias belonging to the active application.

A word-level NNLM holds the majority of parameters in the embedding. Therefore, the disk size for the mixture and AD-NNLM should increase slightly  compared to the baseline architecture. Also the AD-NNLM speed should not increase since it is equivalent to the baseline architecture at inference time.






\section{Experimental setup}
\label{sec:setup}

The training data of our LMs consists of different data sources: anonymized and randomly sampled user requests from both VA and STT that are manually or automatically transcribed, along with synthetic tail-focused datasets. For the latter, we sample from domain-dependent templates  and lists of entities that can fill those slots, both of which are derived from real user data. As mentioned in the introduction, we train NNLMs for three languages: US English, German and Mandarin Chinese.

For our NNLMs, we obtain weights according to the method described in Section~\ref{sec:data}. For the AS models we sample 6B words while for the AA and AD models we sample 12B words.
We run Bayesian hyperparameter optimization and select the final values based on optimal size-accuracy trade off. As a result, the models have a slightly different number of parameters, but we show in section~\ref{sec:results} that this does not impact results noticeably. 
All models have 4 feed-forward layers and an embedding size of 256 -- we tie the input and output embedding weights to reduce disk size~\cite{press}. The hidden size is 768 for the base FOFE model, 512 for the AD FOFE and mixture FOFE and 256 for the Transformer. The Transformer has the same configuration as~\citet{vaswani} and uses 4 attention heads of size 256.
We use the top 100k most frequent words as vocabulary. To speed up training, we use Noise Contrastive Estimation (NCE)~\cite{nce} which is replaced by softmax during inference.

 We train our NNLMs with Block Momentum Stochastic Gradient Descent~\cite{bmsgd}
 with an initial learning rate of 0.256 for AS, AA and AD FOFE and 1.024 for AA Mixture FOFE.
For AS models the optimization converges after 64 epochs while for AA and AD models the optimum is delayed to 128 epochs. 
We keep the initial learning rate fixed for 16 epochs for AS and 64 epochs for the other models and apply a learning rate decay of 0.7 if the heldout perplexity increases for 4 epochs. To stabilize the training a clip norm of 6.0 is applied and the number of NCE samples is set to 4096.

For evaluation, we test on three types of test sets: (1) VA and (2) STT, which consist of user requests sampled according to the distribution that we observe in our VA/STT and thus contain many head queries, and (3) Tail, which is designed to focus on queries with tail entities. Since these do not occur often in our user data, Tail consists of synthetic requests sampled from the same templates and entity lists that generate the synthetic training data.
The requests cover a wide variety of domains such as music, sports and home automation and the audio is generated using Text-to-Speech. 
Table~\ref{tab:test_stats} shows the number of words in each test set.

We evaluate the accuracy of our models using Word Error Rate (WER) and latency using P95 real-time factor (RTF). If $y$ is the duration of the audio signal and $x$ the time it takes to decode $y$, RTF is defined as $x/y$. P95 refers to the 95th percentile and thus captures the latency of the most difficult queries.
We run each test three times and average the RTF numbers to capture outliers.

\begin{table}
\centering
\begin{tabular}{lccc}
\hline
& \textbf{VA} & \textbf{STT} & \textbf{Tail}\\
\hline
English&226k&292k&454k \\
German&130k&154k&204k \\
Mandarin&221k&219k&368k \\
\hline
\end{tabular}
\caption{\label{tab:test_stats}
Number of words per test set per language.
}
\end{table}

The ASR system uses a deep convolutional neural network acoustic model (AM) as described in~\cite{sndcnn,pratap20binterspeech}. For the AS models, we decode the VA and Tail test sets with a VA-specific NNLM and the STT test sets with a STT-specific NNLM. During decoding, the context length of the NNLMs is limited to 8 words to meet the memory and latency contraints of on-device ASR.
We perform a single decoding pass, combining the AM scores with the NNLM scores using optimized weights. We can achieve better WERs by interpolating the NNLM with an n-gram LM trained on tail data and by adding a rescoring pass, but since we want to compare the impact of using different neural architectures, we remove any factors that might obscure that comparison.



\section{Results}
\label{sec:results}

\begin{table}[!h]
\centering
\begin{tabular}{lrrrr}
\hline
\textbf{LM} &\textbf{\#Par}& \textbf{VA} & \textbf{STT} & \textbf{Tail}\\
\hline
\multicolumn{5}{l}{\textbf{English}}\\
\hline
AS FOFE&58M& 4.02 &3.68 &\textbf{17.48} \\
AA FOFE&29M& 4.11 &3.68 &17.78 \\
AA Transf.&27M& \textbf{3.99} & \textbf{3.56} &47.56 \\
AA M-FOFE&37M&\textbf{3.99} &\textbf{3.56} &17.53 \\
AD FOFE&31M&\textbf{3.99} &3.62 &17.51 \\
\hline
\multicolumn{5}{l}{\textbf{German}}\\
\hline
AS FOFE&58M&5.32  & 6.47 &\textbf{29.46} \\
AA FOFE&29M&5.32  &6.35 &29.93 \\
AA Transf.&27M& 11.76  &23.34 &34.42 \\
AA M-FOFE&37M&5.29 &\textbf{6.26} &30.37 \\
AD FOFE&31M&\textbf{5.25} &6.33 &32.36 \\
\hline
\multicolumn{5}{l}{\textbf{Mandarin}}\\
\hline
AS FOFE&58M&5.17&6.04&39.96 \\
AA FOFE&29M&5.25&6.27&38.84 \\
AA Transf.&27M&8.88&13.29&40.66 \\
AA M-FOFE&37M&5.13&\textbf{5.94}&38.16 \\
AD FOFE&31M&\textbf{5.12}&6.05&\textbf{36.41} \\
\hline
\multicolumn{5}{l}{\textbf{Mandarin (equal number of parameters)}}\\
\hline
AS FOFE&68M&5.14&6.00&39.45 \\
AA FOFE&34M&5.26&6.27&38.68 \\
AA Transf.&34M&9.03&13.40&40.38 \\
AA M-FOFE&34M&\textbf{5.10}&6.02&38.54 \\
AD FOFE&34M&5.12&\textbf{5.98}&\textbf{36.48} \\
\hline
\end{tabular}
\caption{\label{tab:wer}
Number of parameters (\#Par) and WERs for the VA, STT and Tail entity test sets for our English, German and Mandarin setups. AS = Application-Specific, AA = Application-Agnostic, AD = Application-Dependent, Transf. = Transformer, M-FOFE = Mixture FOFE.
}
\end{table}

\begin{table}[!h]
\centering
\begin{tabular}{lrrr}
\hline
\textbf{LM} &\textbf{VA} & \textbf{STT} & \textbf{Tail}\\
\hline
\multicolumn{4}{l}{\textbf{English}}\\
\hline
AA Transf. &-18.00  &-21.75  &-11.30  \\
AA M-FOFE  &-23.79  &-31.54  &-17.95  \\
AD FOFE& \textbf{7.40}& \textbf{-8.04}& \textbf{4.66} \\
\hline
\multicolumn{4}{l}{\textbf{German}}\\
\hline
AA Transf. &-19.59  &-13.92  &-24.85  \\
AA M-FOFE &-17.77   &-31.45  &-79.83  \\
AD FOFE & \textbf{7.84}  &\textbf{3.41} &\textbf{5.58}  \\
\hline
\multicolumn{4}{l}{\textbf{Mandarin}}\\
\hline
AA Transf.&-10.06&-14.04&-8.90 \\
AA M-FOFE&-9.89&-30.21&-36.23 \\
AD FOFE&\textbf{-2.11}&\textbf{1.63}&\textbf{-3.83} \\
\hline
\end{tabular}
\caption{\label{tab:rtf}
Latency results: relative P95 RTF reductions with respect to the AA FOFE models for the VA, STT and Tail entity test sets for our English, German and Mandarin setups. AA = Application-Agnostic, AD = Application-Dependent, Transf. = Transformer, M-FOFE = Mixture FOFE.
}
\end{table}

We first evaluate the accuracy of the different neural architectures. Table~\ref{tab:wer} reports the WER for different models on the VA, STT and Tail test sets, along with the number of parameters of the model to give an estimate of the size on disk. Note that for the AS FOFE models, we have twice as many parameters as the AA FOFE models because we train two separate models, one for VA+Tail and one for STT. 

We first observe that moving from AS to AA FOFE and thus reducing the number of parameters by half gives in some cases 1.5-3.8\% WER degradation. Secondly, even though the Transformer architectures have been optimized using Bayesian optimization similar to the FOFE-based models, they give mixed results. For English VA and STT we observe WER improvements while for all other setups we see large degradations. 

The AD FOFE model gives the best accuracy on VA for all languages, while the AA Mixture FOFE gives the best accuracy on STT, but the differences between the two architectures are small. They outperform the baseline AS/AA FOFE and Transformer models in almost all cases. The only exception are the English and German Tail test sets: the AS FOFE models still achieve the best accuracy, probably because infrequent queries benefit the most from doubling the number of parameters. 

As explained in Section~\ref{sec:setup}, we choose hyperparameters based on the optimal accuracy-size trade off. As a result, the number of parameters of the models at the top of Table~\ref{tab:wer} are not exactly the same. 
To ensure that the small size differences do not impact the results significantly, we evaluated results for Mandarin models that all have 34M parameters each and added the results at the bottom of Table~\ref{tab:wer}.
We observe the same trends: the AD FOFE and AA Mixture FOFE give the best results. We confirm that increasing the number of parameters does not lead to better results.


Finally, we report the relative change in P95 RTF (P50 RTF showed the same trend) compared to the baseline AA FOFE model in Table~\ref{tab:rtf}.
Since RTF is hardware-dependent, we mostly care about relative changes compared to the baseline.
We observe that both the Transformer and the Mixture FOFE are significantly slower than the baseline. For the English test sets, the Transformer is faster than the Mixture FOFE, while for German and Mandarin speed depends on the test set.
The AD FOFE gives the fastest inference speed of the proposed models and even outperforms the vanilla FOFE on English VA and all German test sets, while keeping the degradation limited in the other setups.

\section{Conclusion}
\label{sec:concl}

We aim to develop a single NNLM that can serve both VA and STT requests with the same accuracy and speed as application-specific NNLMs, while reducing the disk size approximately by half. We develop a method to optimally balance the data of the VA and STT applications, and propose two novel FOFE feed-forward architectures. The Application-Agnostic Mixture FOFE and the Application-Dependent FOFE both outperform the baseline FOFE and Transformer models in terms of accuracy, and the latter is also competitive in terms of latency.



\section*{Limitations}

The two proposed models (AD FOFE and AA FOFE Mixture) have been tested on more languages than the ones mentioned in this paper, but the comparison with Transformer models has not been done for every language. This paper only uses word-level LMs. We have done preliminary experiments with subword-level LMs but more extensive investigation is needed to draw proper conclusions.

\section*{Ethics Statement}

This paper focuses on the LM of a real-world VA and as such the results cannot be exactly reproduced: we are not aware of any public dataset that mimics our setup, e.g.\ ASR that can serve both VA and STT applications, training data in several languages that exceeds 6B words along with test sets of several hundreds of thousands of words sampled from real user data, etc. 
All data have been anonymized and randomly sampled, and human transcription to create the test sets is only performed from opted-in user data.

\section*{Acknowledgements}

We would like to thank Barry Theobald, Arturo Argueta and Thiago Fraga Da Silva for reviewing this paper.

\bibliography{anthology,custom}
\bibliographystyle{ms}

\appendix

\end{document}